\documentclass[runningheads]{llncs}
\usepackage{graphicx}
\usepackage{tikz}
\usepackage{comment}
\usepackage{amsmath,amssymb} 
\usepackage{color}
\usepackage{hyperref}
\usepackage{biblatex}
\makeatletter
\def\blfootnote{\xdef\@thefnmark{}\@footnotetext}
\makeatother

\usepackage[accsupp]{axessibility}  

\usepackage{breakcites}
\usepackage{adjustbox}
\usepackage[position=top]{subfig}

\begin{document}
\pagestyle{headings}
\mainmatter

\title{Unpaired Image Translation via Vector Symbolic Architectures}

\titlerunning{Unpaired Image Translation via Vector Symbolic Architectures}
%
\author{Justin Theiss\inst{1,2} \and
Jay Leverett\inst{1} \and
Daeil Kim\inst{1} \and
Aayush Prakash\inst{1}}
\authorrunning{J. Theiss et al.}
%
\institute{Meta Reality Labs \\
\email{\{theiss,jayleverett,daeilkim,aayushp\}@fb.com} \\
\and University of California, Berkeley, CA 94720, USA}

\maketitle

\begin{abstract}
Image-to-image translation has played an important role in enabling synthetic data for computer vision. However, if the source and target domains have a large semantic mismatch, existing techniques often suffer from source content corruption aka semantic flipping. To address this problem, we propose a new paradigm for image-to-image translation using Vector Symbolic Architectures (VSA), a theoretical framework which defines algebraic operations in a high-dimensional vector (hypervector) space. We introduce VSA-based constraints on adversarial learning for source-to-target translations by learning a hypervector mapping that inverts the translation to ensure consistency with source content. We show both qualitatively and quantitatively that our method improves over other state-of-the-art techniques.

\keywords{Image-to-image translation, adversarial learning, vector symbolic architectures, semantic flipping.}

\end{abstract}

\blfootnote{\href{https://github.com/facebookresearch/vsait}{https://github.com/facebookresearch/vsait}}

\section{Introduction}
Image-to-image translation techniques~\cite{fu2019geometry,huang2018multimodal,jia2021semantically,lee2018diverse,liu2017unsupervised,park2020contrastive,richter2021enhancing,zhu2017unpaired} have been instrumental in improving synthetic data for computer vision. They have been used to bridge the domain gap for synthetic data~\cite{huang2018multimodal,park2020contrastive,zhu2017unpaired} and for photo-realistic enhancement in virtual reality and gaming applications~\cite{richter2021enhancing}. Some researchers~\cite{devaranjan2020meta,kar2019meta,prakash2021self} argue that the domain gap can be further factorized into a content (shift in semantic statistics) and appearance gap.
In this work, we are interested in the unpaired image-to-image translation method in the challenging scenario where the content gap between the source and target domains is large.

\begin{figure*}[t!]
    \centering
    \includegraphics[width=\hsize]{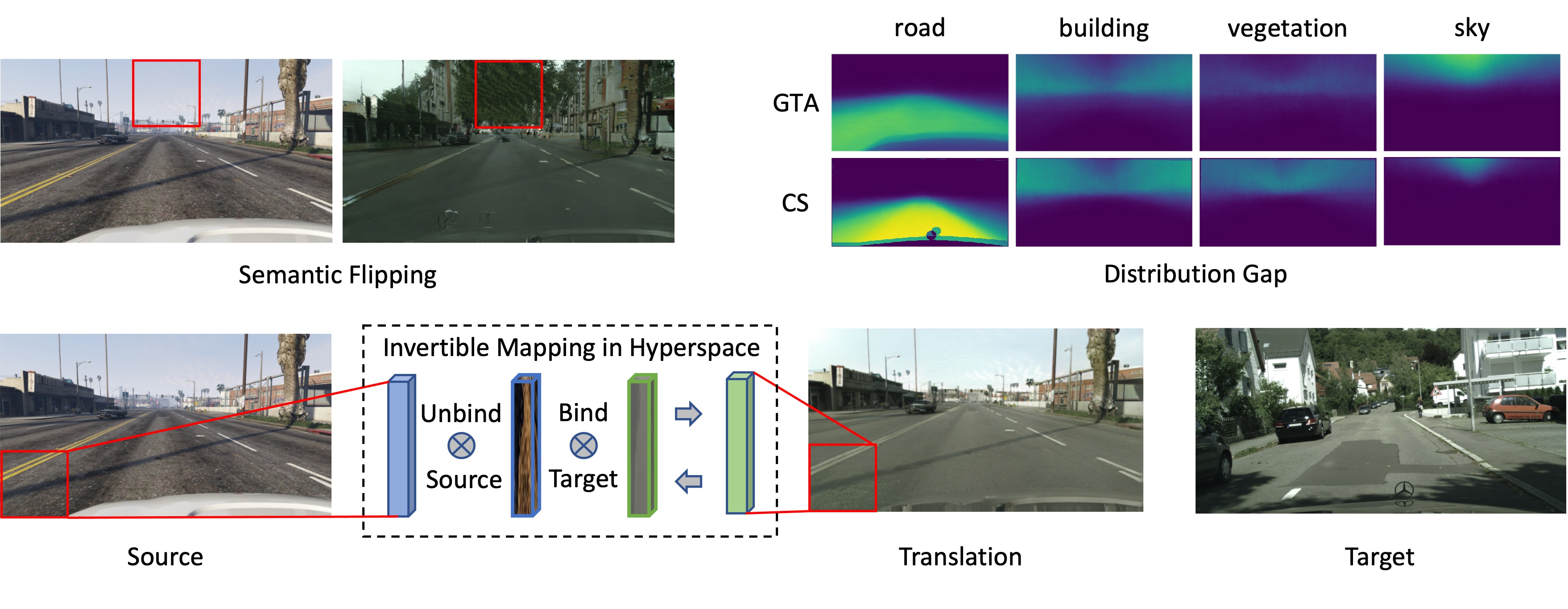}
    \caption{We propose Vector Symbolic Architecture based image-to-image translation technique (VSAIT) which addresses semantic flipping (source content corruption) that happens when the distribution gap (shift in semantic statistics) between source and target domains is large. Our method learns a mapping in high-dimensional vector space (hyperspace), which encourages translated images to be consistent with the source domain. Conceptually, our approach aims to ``unbind'' source-related information (e.g., texture and color) and ``bind'' target-related information as well as vice versa to recover source content.}
    \label{fig:teaser}
    \vspace{-1 em}
\end{figure*}

There are several techniques for unpaired image-to-image translation~\cite{huang2018multimodal,park2020contrastive,zhu2017unpaired}. Some approaches assume that content and style can be separated in order to translate style without corrupting content (e.g., \cite{huang2018multimodal}). Others have used a bijective mapping to reconstruct the source image from the translated image (i.e., a ``cyclic loss'' \cite{zhu2017unpaired}). 
However, these methods do not work well if there is a \emph{large shift in distribution} between source and target domains leading to the problem of \emph{semantic flipping}. Semantic flipping is characterized by image artifacts, object or feature hallucinations that are a result of adversarial training for datasets with a large content gap. Specifically, this is observed as a change in content between source and translated images (e.g., sky to trees in Figure \ref{fig:teaser}). Semantic flipping is a critical issue for improving photorealism in computer graphics applications \cite{richter2021enhancing} as well as training downstream tasks using translated images, as we want to preserve the source semantic labels of translations.

Relatively few works have directly focused on the semantic flipping problem; however, they can be broadly categorized into three approaches: image-level consistency, domain-invariant encoding, and task-level consistency. Methods using an image-level consistency loss attempt to ensure that pixel-level information is highly correlated between source and translated images \cite{benaim2017one,fu2019geometry}. Such methods may fail to account for feature-level differences that do not correlate well with pixel-level differences. Approaches that focus on domain-invariant encoding train the generator to encode domain-invariant content either using a shared latent space \cite{lee2018diverse} or contrastive learning \cite{jia2021semantically}. These methods will fail to reduce semantic flipping if content and style cannot be sufficiently disentangled. Finally, others have used pre-trained task networks to generate pseudo-labels for each domain to ensure a consistent translation with respect to source labels (i.e., semantic masks) \cite{hoffman2018cycada,richter2021enhancing}. However, these methods fail if the task network cannot generate pseudo-labels for both domains. We instead focus on a method that provides feature-level consistency between source and translated images without explicit assumptions of domain-invariant encoding or quality of pseudo-labels. This addresses gaps in previous methods which make them vulnerable to semantic flipping.

In the current paper, we propose a novel usage of a theoretical framework known as vector symbolic architectures (VSA~\cite{gayler2004vector,kanerva2009hyperdimensional}; also referred to as hyperdimensional computing) as a new paradigm for unpaired image-to-image translation. Although much of the VSA research has been conducted in theoretical neuroscience (see \cite{kleyko2021survey2,kleyko2021survey1}), it has more recently been applied to computer vision problems using extracted features from deep neural networks \cite{neubert2021hyperdimensional,neubert2019introduction}. VSA defines a high-dimensional vector (hypervector) space and operators used for symbolic computation, which allows for mathematical formulations of conceptual queries such as, ``what color is the car?''. In the case of unpaired image translation, such formulations are useful because they enable us to recover attributes from the source image and ensure consistent relationships among features when translating images from one domain to another. VSA is well suited for this approach as it can represent arbitrary symbols and formulations generally without supervision or training \cite{purdy2016encoding}. The important difference from previous methods addressing semantic flipping is that VSA ensures that hypervector representations of different semantic content are almost orthogonal to each other (e.g., sky and trees)~\cite{kanerva2009hyperdimensional}. The cosine distance between translated and source hypervectors therefore is greatest when semantic flipping occurs.

Using this framework, we propose a method for learning a hypervector mapping between source and target domains that is robust to semantic flipping (Figure \ref{fig:teaser}). By inverting this mapping, we are able to minimize the distance between features extracted from source and translated images without requiring that content and style be fully disentangled. We demonstrate qualitatively and quantitatively that this approach significantly reduces the image artifacts and hallucinations observed for unpaired image translation between domains with a large content gap. We hope that our work provides inspiration for incorporating VSA into new areas of computer vision research. Our contributions include:

\begin{itemize}
    \item Our method addresses important artifacts and feature hallucinations (semantic flipping) that often occur with other unsupervised image translation methods as a result of content gap between source and target domains.
    \item To the best of our knowledge, we are the first to show that Vector Symbolic Architectures (VSA) can be used for a challenging task of image translations in the wild. We hope this opens up an exciting area of research around VSA. 
    \item We demonstrate qualitative and quantitative improvement over state-of-the-art methods that directly address semantic flipping across multiple experiments with unmatched semantic statistics.
\end{itemize}

\section{Related Work}
\subsubsection{Unpaired Image-to-Image Translation}

Unpaired image-to-image translation is a widely studied problem with many existing techniques~\cite{almahairi2018augmented,benaim2017one,fu2019geometry,huang2018multimodal,isola2017image,jia2021semantically,kim2017learning,lee2018diverse,liu2017unsupervised,park2020contrastive,richter2021enhancing,yi2017dualgan,zhu2017unpaired}. One popular approach is to impose the cycle-consistency constraint~\cite{kim2017learning,yi2017dualgan,zhu2017unpaired}, which states that a source-to-target generator and a target-to-source generator are bijections. Building on cycle-consistency, UNIT~\cite{liu2017unsupervised} mapped the source and target domains to a shared latent space, while DRIT~\cite{lee2018diverse} and other methods~\cite{almahairi2018augmented,huang2018multimodal} factorized the latent space into a shared content space and a domain-specific style space. DistanceGAN~\cite{benaim2017one} showed that source to target translations can be learnt unidirectionally. CUT~\cite{park2020contrastive} subsequently proposed a unidirectional method based on contrastive learning. Whereas some of these methods require separate networks per domain or assume that content and style can be disentangled, our method uses a single generator without assumption of disentanglement.

\begin{figure*}[t!]
    \centering
    \includegraphics[width=\hsize]{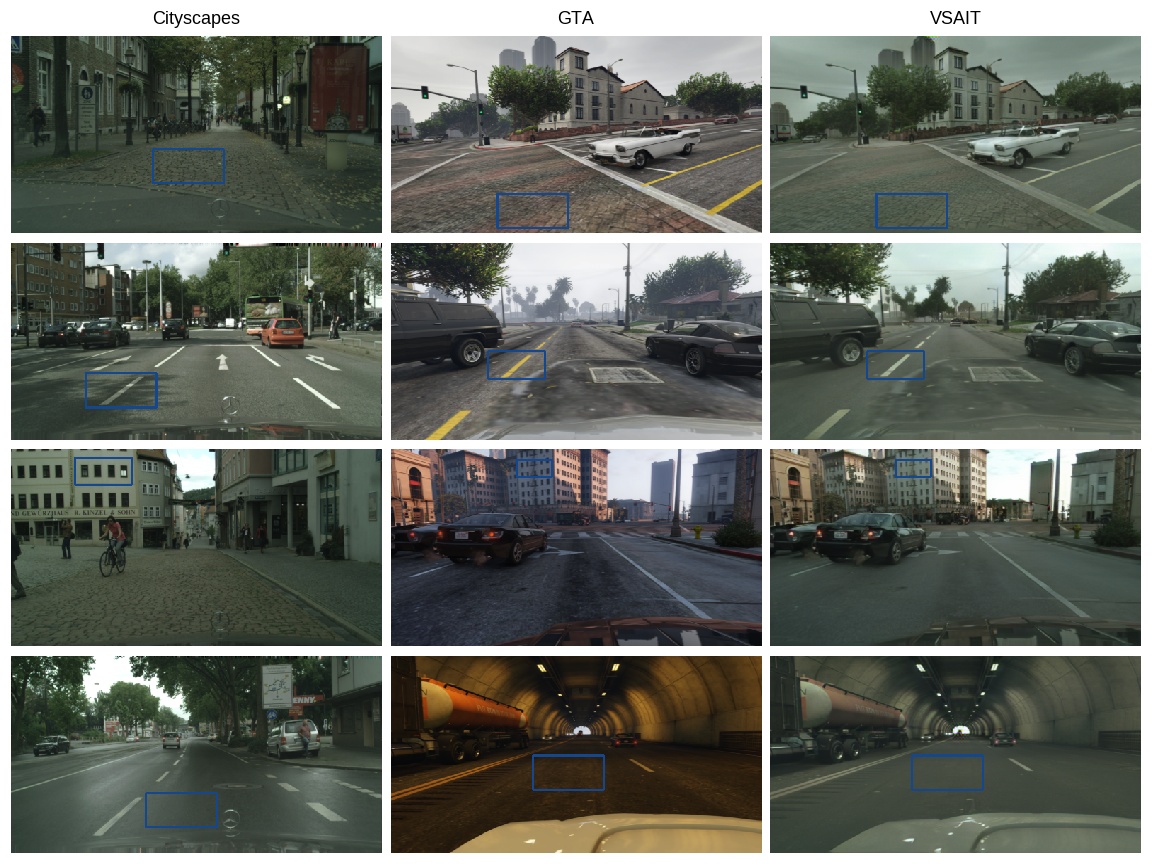}
    \caption{Example GTA translations from our method (VSAIT) alongside representative examples from Cityscapes. We can see that VSAIT successfully translates GTA images to Cityscapes' style and that, in particular, our method is able to learn specific textures and attributes (e.g. gray cobblestone, white lane lines rather than yellow) that are commonly found in Cityscapes images. Moreover, VSAIT is able to handle unseen scenarios (e.g. tunnel) that do not occur in Cityscapes without semantic flipping. }
    \label{fig:cityscapes_vs_vsait}
\end{figure*}

\subsubsection{Semantic Flipping}
The body of research on semantic flipping is relatively small. Methods based on cycle-consistency combat semantic flipping through pixel-level reconstruction, but they are unable to prevent semantic flipping in the face of large distributional shifts. GcGAN~\cite{fu2019geometry} enforces geometry-consistent constraints on image translations by ensuring robustness for transformations such as flipping or rotation. This does not necessarily address semantic flipping for small, local features that could still be geometry-consistent. EPE~\cite{richter2021enhancing} addresses semantic flipping by conditioning on G-buffers (e.g. albedo, glossiness, depth, normals) as well as incorporating semantic segmentation pseudo-labels into the discriminator. Although their method addresses semantic flipping, it does not generalize to many tasks or datasets since it assumes access to the synthetic image rendering process and a downstream task network that can generate pseudo-labels for both source and target domains. Alternatively, SRUNIT~\cite{jia2021semantically} proposed to address semantic flipping by using contrastive learning (following CUT~\cite{park2020contrastive}) to encode domain-invariant semantic representations that are robust to small feature-space perturbations. Although their approach does not require access to any labels or rendering process like EPE, it does not sufficiently address semantic flipping. Unlike these approaches, we reduce semantic flipping by inverting the translation in hyperspace using VSA in order to ensure recovery of source information. 

\subsubsection{Computer Vision Applications of VSA}
Although much of the research associated with VSA has been conducted in theoretical neuroscience, Neubert et al. \cite{neubert2019introduction} provided examples of how VSA can be used for computer vision tasks such as object and place recognition, which demonstrated how image features extracted from pre-trained neural networks can be used within the VSA framework. Other works have used VSA for visual question answering \cite{montone2017hyper} and scene transformation \cite{kent2017vector}, albeit with simple shapes or MNIST digits. More recently, Osipov et al. \cite{osipov2021hyperseed} used VSA to learn to unbind category representations among unlabelled data vectors bound to a shared representational space. This is most similar to the challenge that we are addressing in the current paper, as we can consider the source and target features to be bound to a similar content space; however, unlike their study we do not have access to the underlying shared representational space.

\section{Method}

\subsection{Preliminary: VSA}
\label{sc:prelim}
Vector Symbolic Architectures (VSA) \cite{gayler2004vector,kanerva2009hyperdimensional} provide a framework for encoding and manipulating information in a hypervector space (hyperspace) with high capacity and robustness to noise \cite{kanerva2009hyperdimensional}. 
In VSA framework, hypervectors are randomly sampled from a vector space $\mathbb{V} = [-1, 1]^{n}$ (where $n >> 1000$) in order to represent symbols that can be typically manipulated using two operators --  binding/unbinding (element-wise multiplication) and bundling (element-wise addition). The binding operator can be used to assign an attribute to a symbol (e.g., ``gray car''), whereas bundling can be used to group symbols together (e.g., ``gray car and red bike''). These operators define the Multiplication-Addition-Permutation (MAP) architecture. Other operators and architectures (see \cite{schlegel2021comparison}) are beyond the scope of our work.

The benefit of using VSA framework for image translation is its versatility in representing arbitrary symbols without supervision. This allows us to infer underlying representational spaces (e.g., content distribution) without explicitly defining them. Since the challenge of semantic flipping in unpaired image translation is typically to constrain the generator, we do not need to learn how to generate images from hypervectors but instead learn a mapping that ensures we recover the same source information for a given translated hypervector.

In order to assign random hypervectors to image features, Neubert et al.~\cite{neubert2019introduction} used locality sensitive hashing (LSH) to project image features extracted using a pre-trained neural network (AlexNet \cite{krizhevsky2012imagenet}) to a relatively lower-dimensional space (from $13 \times 13 \times 384$ to $8192$). In this case, the entire image was encoded into a single hypervector; however, image patches can also be encoded as separate hypervectors \cite{neubert2021hyperdimensional}.
As an example, imagine an image patch from the source domain that contains a car and pedestrian. We can assume without loss of generality that the image patch is represented as a hypervector:

\begin{equation}
    v_{src} = c \otimes c_{src} + p \otimes p_{src},
    \label{eqn:v_src_example}
\end{equation}
\noindent where $c$ and $p$ are hypervectors representing the car and pedestrian bound with source-specific hypervectors $c_{src}$ and $p_{src}$, respectively. 

Since we do not have access to the underlying symbols associated with source and target domains, instead we must learn a mapping that unbinds source-specific information and binds target-specific information (and vice versa). 
Specifically, we would like to learn a hypervector mapping that can unbind these source-specific attributes and bind target-specific attributes in order to obtain a target representation $v_{tgt}$ as shown in Equation \ref{eqn:mapping_example}.

\begin{equation}
    \begin{aligned}
        & v_{src} \otimes u_{src \leftrightarrow tgt} \approx v_{tgt}, \\
        & \text{where } u_{src \leftrightarrow tgt} = (c_{src} \otimes c_{tgt} + p_{src} \otimes p_{tgt})
    \end{aligned}
    \label{eqn:mapping_example}
\end{equation}

Since the binding/unbinding operation is invertible, $c_{src}$ and $p_{src}$ are unbound while $c_{tgt}$ and $p_{tgt}$ are correspondingly bound. Importantly, since these random hypervectors are very likely to be almost orthogonal, binding/unbinding the incorrect attributes (i.e., $c \otimes p_{src}$) simply adds noise to the resulting vector as demonstrated in Equation \ref{eqn:vsa_algebra}. 

\begin{equation}
    \begin{aligned}
        v_{src} \otimes u_{src \leftrightarrow tgt} & = 
         c \otimes c_{src} \otimes (c_{src} \otimes c_{tgt} + p_{src} \otimes p_{tgt}) \\
        & + p \otimes p_{src} \otimes (c_{src} \otimes c_{tgt} + p_{src} \otimes p_{tgt}) \\
        & = (c \otimes c_{tgt} + noise) + (noise + p \otimes p_{tgt}) \\
        & = c \otimes c_{tgt} + p \otimes p_{tgt} + noise 
         \approx v_{tgt}
    \end{aligned}
    \label{eqn:vsa_algebra}
\end{equation}

It is also worth noting that $u_{src \leftrightarrow tgt}$ is equivalent to $u_{tgt \leftrightarrow src}$, which means a single vector can be used to map between the domains. Hence, $ v_{tgt} \otimes u_{src \leftrightarrow tgt} \approx v_{src}$. Note that hypervectors in the VSA framework are distributed representations that are very robust to noise \cite{ahmad2015properties} and further that cosine similarity will still be high relative to similarity with random vectors. The above example demonstrates the unique properties of the VSA framework, which allow for algebraic formulations of symbolic representations.

\begin{figure*}[t!]
    \centering
    \includegraphics[width=\hsize]{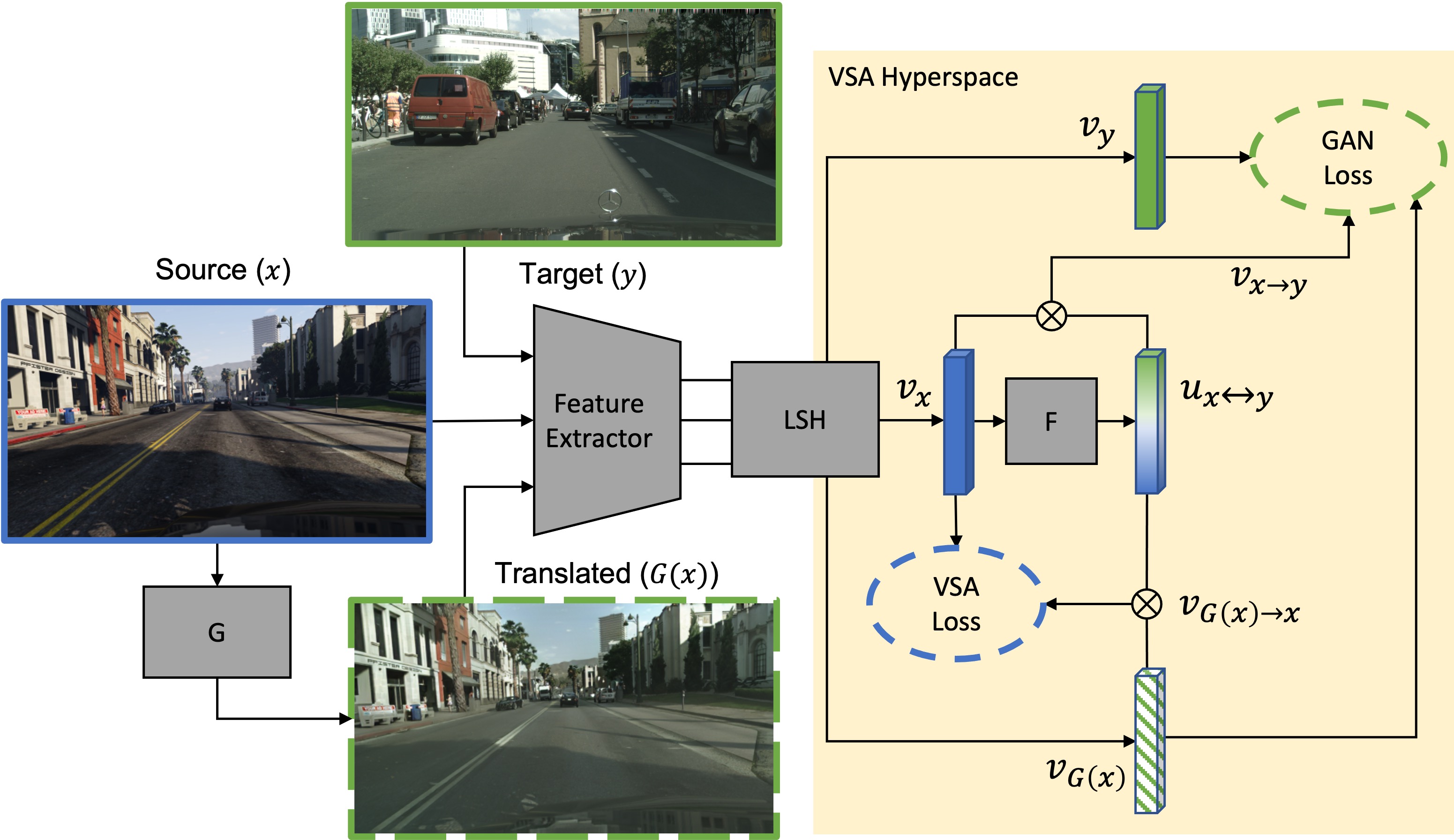}
    \caption{Our method addresses semantic flipping by learning an invertible mapping in a high-dimensional vector space (VSA hyperspace) in order to ensure consistency between source and translated images. We extract features and use locality sensitive hashing (LSH) to encode source ($x$, blue), target ($y$, green), and translated ($G(x)$, striped green) images into this random VSA hyperspace. We use a GAN loss to train $G$ to generate images with hypervectors similar to those in the target domain. Finally, the hypervector mapping ($u_{x \leftrightarrow y}$, green-and-blue vertical bar) is used to invert our translation (Eqn \ref{eqn:mapping_vector}) to recover the source hypervectors (VSA Loss). See Section~\ref{subsc:training} for more details.}
    \label{fig:method}
\end{figure*}

\subsection{VSA-Based Image Translation (VSAIT)}
\label{subsc:vsait}
\subsubsection{Overview}
Our goal is to translate images from a source to target domain, while minimizing semantic flipping caused by differences in the content distribution between the two domains. As shown in Figure \ref{fig:method}, our method uses a hypervector adversarial loss (GAN Loss in Figure \ref{fig:method}) that operates on VSA-based hypervector representations of image patches. In order to address semantic flipping we constrain the generator to preserve the content of the source domain via a VSA-based cyclic loss (VSA Loss in Figure \ref{fig:method}). 
We do so by leveraging VSA to invert our translation in hypervector space to ensure consistency with the source content. We will cover these two loss functions later in this section. 
Our method has three major network components-- Generator $G$, Discriminator $D_{Y}$ and Mapper $F$ that we will discuss next. We present training details in Section~\ref{subsc:training}. 

\subsubsection{Generator}
Our approach uses a generator $G : X \to Y$ (shown in Figure \ref{fig:method}) and feature-level discriminator $D_{Y}$ in order to learn an unpaired image translation between $X$ (source) and $Y$ (target) domains, where $D_{Y}$ is trained to discriminate between hypervectors extracted from target and translated images. 

As previously introduced in Section~\ref{sc:prelim}, to assign hypervectors to image patches, we follow Neubert et al.~\cite{neubert2019introduction} and extract features using a pre-trained neural network (in our case, concatenated multiple layers of VGG19 \cite{simonyan2014very}). These concatenated features are then randomly projected from the extracted feature vector $f_i \in \mathbb{R}^{m}$ to the random vector space $\mathbb{V} = [-1, 1]^{n}$ where $m >> n$. 
Using this approach, we denote hypervectors extracted from target images as $v_{Y}$, source images as $v_{X}$, and translated images as $v_{G(X)}$ (see Figure \ref{fig:method}).

\subsubsection{Source $\leftrightarrow$ Target Mapper}
Furthermore, we train a network $F$ to generate a hypervector mapping $u_{X \leftrightarrow Y}$ (e.g., Eqn~\ref{eqn:vsa_algebra}) between source and target domains. We use this hypervector mapping to invert our translation, providing a VSA-based cyclic loss in hyperspace.
The invertible mapping accomplishes two operations with a single step: unbinding of source (resp. target) representations and binding of target (resp. source) representations. This means that if we apply this mapping to our translated hypervectors $v_{G(X)}$, we should approximately obtain the original source hypervectors $v_{X}$. Similarly, if we apply this mapping to source hypervectors $v_{X}$, we approximately obtain translated hypervectors $v_{G(X)}$. We denote translated hypervectors that have been mapped to the source domain as $v_{G(X) \to X}$ and those mapped from source to target domain as $v_{X \to Y}$ as shown in Equation \ref{eqn:mapping_vector} as well as Figure \ref{fig:method}.

\begin{equation}
    \begin{aligned}
        v_{G(X)} \otimes u_{X \leftrightarrow Y} & = v_{G(X) \to X} \approx v_{X} \\
        v_{X} \otimes u_{X \leftrightarrow Y} & = v_{X \to Y} \approx v_{G(X)}
    \end{aligned}
    \label{eqn:mapping_vector}
\end{equation}

\subsubsection{Hypervector Adversarial Loss}
In order to train $G$,  $D_{Y}$ and $F$ to translate images that match the target distribution, we use an adversarial loss \cite{goodfellow2014generative}. Specifically, we want the hypervectors of our translated image $v_{G(X)}$ to be similar to those from the target domain $v_{Y}$. Furthermore, by binding the hypervector mapping $u_{X \leftrightarrow Y}$ to source vectors $v_{X}$, we should obtain a hypervector of the source features mapped to the target domain (Eqn~\ref{eqn:mapping_vector}) . Therefore, we train $F$ along with $G$ and $D_{Y}$ using the following adversarial loss:

\begin{equation}
    \begin{aligned}
        \mathcal{L}_{GAN}(G, D_{Y}, F, X, Y) = & \mathbb{E}_{y \sim p_{Y}(y)}[\text{log} D_{Y}(v_{y})] \\ 
        + & \mathbb{E}_{x \sim p_{X}(x)}[\text{log}(1 - D_{Y}(v_{G(x)})] \\
        + & \mathbb{E}_{x \sim p_{X}(x)}[\text{log}(1 - D_{Y}(v_{x} \otimes F(v_{x}))],
    \end{aligned}
    \label{eqn:gan_loss}
\end{equation}

\noindent where $G$ and $F$ networks are trained to fool the discriminator $D_{Y}$, thereby minimizing the objective while $D_{Y}$ attempts to maximize it.

\subsubsection{VSA-Based Cyclic Loss}
Although adversarial training provides a method to generate images that match the target distribution, the differences in content between domains will result in semantic flipping. We therefore need a loss that constrains the generator to preserve source content and reduce semantic flipping. To do so, we incorporate our VSA-based cyclic loss, which ensures that the same hypervectors are obtained when mapping translated vectors back from target to source domain ($v_{G(X) \to X}$):

\begin{equation}
    \mathcal{L}_{VSA}(G, X) = \mathbb{E}_{x \sim p_{X}(x)} \left[\frac{1}{n} \sum_{i=1}^{n} dist\left(v^{i}_{x}, v^{i}_{G(x) \to x}\right)\right],
    \label{eqn:vsa_loss}
\end{equation}

\noindent where $dist(\cdot, \cdot)$ is the cosine distance averaged across the $n$ hypervectors (representing image patches indexed by $i$) for the image $x$. By minimizing the cosine distance ($1 - \text{cosine similarity}$) between $v_{G(X) \to X}$ and $v_{X}$, we ensure that the same features are recovered after inverting our translation (i.e., representations of ``car'' and ``pedestrian'' are maintained in the example from Equation \ref{eqn:v_src_example}). 

Our overall objective combines our adversarial and VSA-based cyclic losses, training $G$ to generate images matching the target domain and $F$ to invert the translation to ensure consistency with the source domain:

\begin{equation}
    \mathcal{L}(G, D_{Y}, F, X, Y) = \mathcal{L}_{GAN}(G, D_{Y}, F, X, Y) + \lambda \mathcal{L}_{VSA}(G, X),
    \label{eqn:total_loss}
\end{equation}

\noindent where $\lambda$ controls the relative importance of our VSA-based cyclic loss.

\subsection {VSAIT Training}
\label{subsc:training}

We train our method with a GAN-based training framework using the objective described in Equation~\ref{eqn:total_loss}. We begin by randomly sampling an image from each domain and translating the source image into the target domain via the generator $G$. 
In order to use the VSA framework, we must first encode data into the hypervector space. Therefore, we extract features from each image (source, target, and translated) at multiple layers of a pre-trained neural network (Feature Extractor in Figure \ref{fig:method}) as the first step in encoding image patches into the VSA hyperspace. We consider each image patch to be represented by the extracted features sharing its receptive field. We flatten and concatenate these features across layers, resulting in a set of feature vectors (one per image patch) with a dimensionality of $m$.

We then use LSH (Figure \ref{fig:method}) to reduce the dimensionality of the extracted feature vectors into our random hyperspace $\mathbb{V}$, which reduces the dimensionality to $n << m$. Specifically, we normalize and project the extracted feature vectors using a random matrix where each row is sampled from a standard normal distribution and normalized to unit length. The resulting vectors are therefore in the range $[-1, 1]$, which is necessary to implement the VSA binding operation as described in Section \ref{sc:prelim}. These hypervectors represent the image patches for the source, target, and translated images as shown in Figure \ref{fig:method}. Note that this process does not require training nor is it necessarily dependent on a specific feature extractor, although different feature extractors may encode different information. 

Finally, we use $F$ to generate a hypervector mapping for each source hypervector. We use these hypervector mappings ($u_{x \leftrightarrow y}$ in Figure \ref{fig:method}) to unbind source (resp. target) information and correspondingly bind target (resp. source) information (Eqn \ref{eqn:mapping_vector}). By applying this mapping to the source hypervectors, we should obtain a hypervector representation of the source image translated into the target domain ($v_{x \to y}$ in Figure \ref{fig:method}), which is used in our adversarial loss (Eqn~\ref{eqn:gan_loss}) to train $F$. The ultimate goal of this step, however, is to use the mapping to invert our translation $v_{G(x)}$ and ensure that we recover the same source information $v_{x}$ (Eqn~\ref{eqn:vsa_loss}). Doing so will reduce semantic flipping by constraining $G$ to generate images that have hypervectors that are consistent with the source domain.

\section{Experiments}
We evaluate our method using experiments across multiple datasets. First, we compare against baseline techniques for GTA \cite{richter2016playing} to Cityscapes \cite{Cordts_2016_CVPR_cityscapes}, where the two domains have inherent semantic differences. We then perform experiments using datasets sub-sampled to create differences in semantic statistics. Specifically, we follow the method in \cite{jia2021semantically} to sub-sample the Google Maps \cite{isola2017image} dataset, which is typically designed for paired image translation tasks. We show that VSAIT works as intended to reduce semantic flipping while still generating diverse image translations. Our qualitative results in Figure \ref{fig:cityscapes_vs_vsait} demonstrate that the hyperspace mapping constrains the generator to translate features that are consistent between source and target domains. Furthermore, we show that other methods still have significant artifacts and hallucinations related to semantic flipping (Figure \ref{fig:comparison1}). Our quantitative results show improvements against the other baselines, particularly for the GTA to Cityscapes task. Finally, we perform an ablation study to evaluate the contributions of different components in VSAIT.

\subsection{Implementation Details}
\label{subsc:implementation}

We follow CUT \cite{park2020contrastive} in our choice of the generator network architecture. For the discriminator network, we use a three-layer fully-convolutional network with $1 \times 1$ convolutional filters. For the mapping network $F$, we use a two-layer fully-convolutional network. We train VSAIT using the Adam optimizer \cite{kingma2015adam} ($\beta_{1} = 0$, $\beta_{2} = 0.9$) with a batch size of 1 and learning rate of $0.0001$ for the generator (and mapping network $F$) and $0.0004$ for the discriminator. To improve adversarial training, we use the ``hinge loss'' \cite{lim2017geometric} rather than the negative log-likelihood objective in $\mathcal{L}_{GAN}$ (Eqn \ref{eqn:gan_loss}). For GTA to Cityscapes, we use $\lambda = 10$ in Equation \ref{eqn:total_loss} and $\lambda = 5$ for other experiments. See Supplemental for more details.

\subsection{Datasets}
We perform our experiments using three datasets: GTA \cite{richter2016playing}, Cityscapes \cite{Cordts_2016_CVPR_cityscapes}, and Google Maps \cite{isola2017image}. GTA \cite{richter2016playing} is a synthetic dataset of 24966 images generated from the video game Grand Theft Auto V. Cityscapes \cite{Cordts_2016_CVPR_cityscapes} is a real dataset of driving scenarios accompanied with finely-annotated semantic segmentation labels, which includes 2975 training and 500 validation images. The Google Maps dataset \cite{isola2017image} is a collection of 2194 paired maps and aerial photos (1096 training and 1098 validation images).

\subsection{Baselines}
We compare our method against several baselines that are relevant to the semantic flipping problem: GcGAN, DRIT, CUT, SRUNIT, and EPE. We choose these methods for comparison as they reflect the recent approaches focused on reducing semantic flipping. Specifically, they represent methods that use image-level consistency (GcGAN \cite{fu2019geometry}), domain-invariant encoding (DRIT \cite{lee2018diverse}, CUT \cite{park2020contrastive}, SRUNIT \cite{jia2021semantically}), and task-level consistency (EPE \cite{richter2021enhancing}). It is worth reiterating that EPE only works for synthetic datasets with access to G-buffers (e.g., albedo, glossiness, depth, normals), which are not publicly available for GTA and do not exist for real datasets (e.g., Google Maps). Therefore we compare qualitatively to EPE and rely on quantitative values reported in their work.

\subsection{GTA $\to$ Cityscapes}
\begin{figure*}[t!]
    \centering
    \includegraphics[width=\hsize]{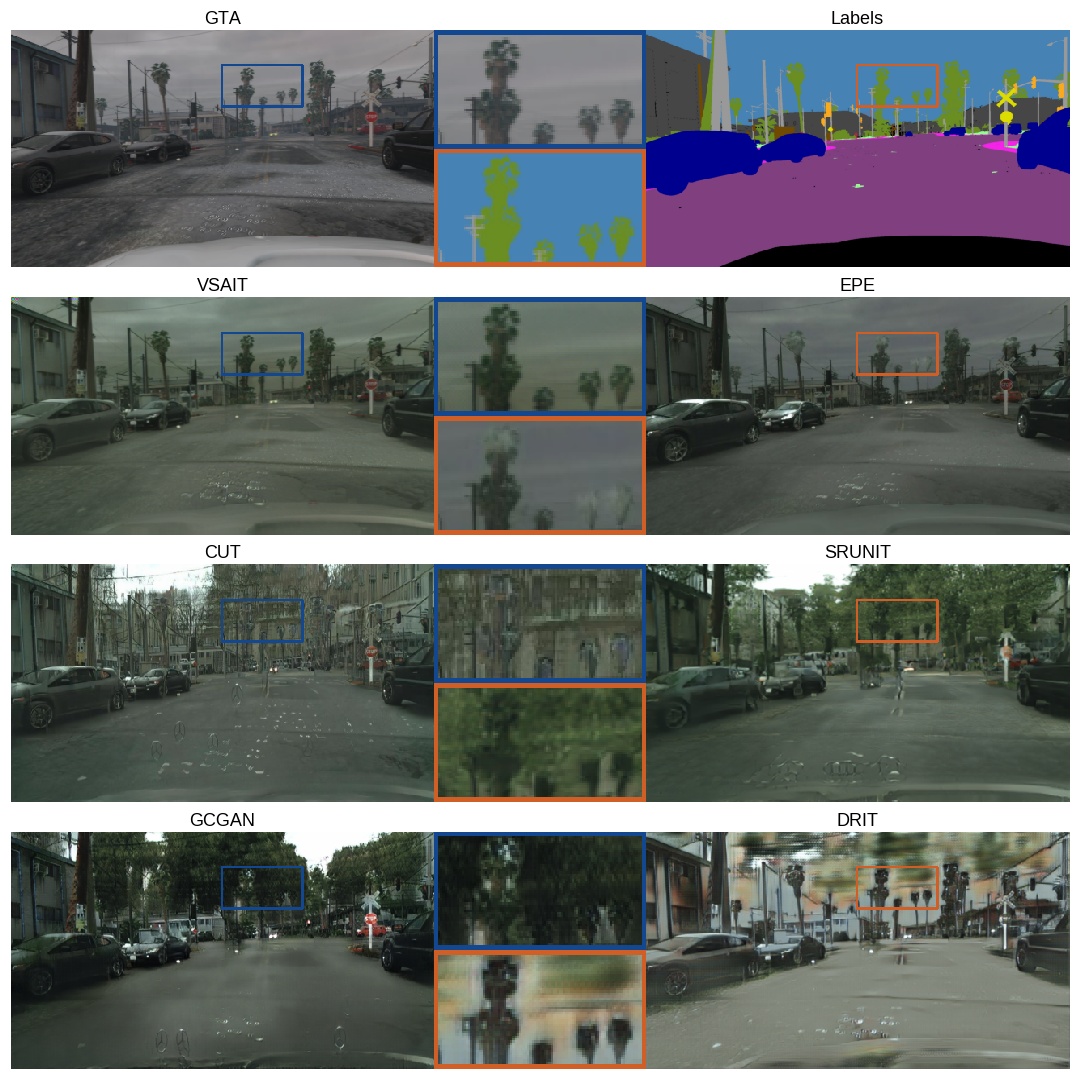}
    \caption{We compare our VSAIT translation to those of baseline methods on an example GTA image. Typical methods often suffer from semantic flipping in open sky regions of GTA, as such regions are observed less often in Cityscapes (i.e., content gap). Here we see that SRUNIT \cite{jia2021semantically}, CUT \cite{park2020contrastive}, GcGAN \cite{fu2019geometry} and DRIT \cite{lee2018diverse} each suffer from sky hallucinations, while EPE \cite{richter2021enhancing}  removes palm trees (features that are absent from Cityscapes) in the distance. Meanwhile, our method is able to translate to Cityscapes style while preserving  semantics from the original GTA image.}
    \label{fig:comparison1}
\end{figure*}

We first demonstrate our method using GTA to Cityscapes, which are unpaired datasets that naturally differ in their semantic statistics. For training, we use 20000 of GTA images for our source dataset and all 2975 training images for our target dataset. Following \cite{richter2021enhancing}, we evaluate our image translation performance using the Kernel Inception Distance (KID) \cite{binkowski2018demystifying} which measures the distance of extracted features from translated and target images using the pre-trained InceptionV3 network \cite{szegedy2016rethinking}. As seen in the Table \ref{tab:ablation_results}, our method outperforms the baseline methods in the KID metric. As shown in Figure \ref{fig:comparison1}, only EPE has similar quality of image translations for GTA to Cityscapes. However, whereas other methods hallucinate trees and Mercedes hood ornaments, EPE has a tendency to remove palm trees and add unnatural lighting to cars even in dark scenes.

We additionally evaluate translation quality via semantic segmentation metrics computed using DeepLab V3 \cite{chen2017rethinking} pretrained on Cityscapes.
In line with \cite{jia2021semantically}, we report three metrics related to semantic segmentation performance (Table \ref{tab:unmatched_results}): pixel accuracy, class accuracy, and mean intersection over union (mIoU). As shown in Table \ref{tab:unmatched_results}, we outperform the other baselines by a wide margin. Whereas KID computes the distance between translated and target images in feature space, semantic segmentation metrics better reflect semantic flipping directly. As seen in Figure \ref{fig:comparison1}, other methods often generate trees and other features in the sky, which will lower semantic segmentation performance in those regions. 

\begin{table}[h!]
    \centering
    \caption{Quantitative evaluation across datasets with unmatched semantics. The metrics included are average pixel prediction accuracy (pxAcc), average class prediction accuracy (clsAcc), mean IoU (mIoU), average L2 distance (Dist), and pixel accuracy with task-specific thresholds (Acc).}
    \begin{adjustbox}{max width=\textwidth}
    \begin{tabular}{c|ccc*{2}{|ccc}}
        \hline
         & \multicolumn{3}{c|}{GTA $\to$ Cityscapes} & \multicolumn{3}{c|}{Map $\to$ Photo} & \multicolumn{3}{c}{Photo $\to$ Map} \\
        \hline
        Method  & pxAcc & clsAcc & mIoU & Dist & Acc($\delta_{1}$) & Acc($\delta_{2}$) & Dist & Acc($\delta_{1}$) & Acc($\delta_{2}$) \\
        \hline
        GcGAN~\cite{fu2019geometry} &  65.62 & 32.38 & 22.64 & 71.47 & 28.87 & 43.48 & 23.62 & 15.00 & 30.65 \\
        DRIT~\cite{lee2018diverse} &  64.28 & 32.17 & 20.99 & 70.87 & 28.97 & 43.56 & 24.19 & 13.94 & 29.01 \\
        CUT~\cite{park2020contrastive} &  64.59 & 32.19 & 20.35 & 70.28 & 28.86 & 44.07 & 23.44 & 16.25 & 31.34 \\
        SRUNIT~\cite{jia2021semantically} &  67.21 & 32.97 & 22.69 & 68.55 & 30.41 & 45.91 & 23.00 & \textbf{17.67} & \textbf{32.78} \\
        VSAIT (ours) &  \textbf{76.48} & \textbf{45.33} & \textbf{30.89} & \textbf{64.98} & \textbf{45.3} & \textbf{69.28} & \textbf{22.86} & 15.2 & 32.13 \\
        \hline
    \end{tabular}
    \end{adjustbox}
    \label{tab:unmatched_results}
\end{table}

\subsection{Google Map $\to$ Aerial Photo}
\label{subsc:map2photo}
We then demonstrate performance on the Google Maps \cite{isola2017image} dataset. Here we sub-sample the 1096 training images using K-means clustering of the histograms from grayscale map images to obtain two datasets with different semantic statistics (as in \cite{jia2021semantically}). We evaluate translation performance on all 1098 validation images and report three pixel-level metrics (Table \ref{tab:unmatched_results}). The first metric is the L2 distance between translated and target images. We also report pixel accuracy defined as the percentage of pixels with a maximum absolute difference less than a given threshold (i.e., $max(|r_i - r_i'|, |g_i - g_i'|, |b_i - b_i'|) < \delta$). For the map to aerial photo task, we use thresholds of $\delta = 30, 50$ \cite{jia2021semantically}. Similar to GTA to Cityscapes, our approach substantially outperforms the baseline methods. Particularly interesting is the improvement we see in the pixel accuracy for both thresholds ($\delta = 30, 50$), whereas the other methods perform very similarly.

\subsection{Aerial Photo $\to$ Google Map}
We additionally demonstrate performance for the task of photo to map using the same sub-sampled datasets obtained from the Google Maps training dataset as described above. For this experiment, we evaluate the same metrics as with map to photo but with pixel accuracy threshold of $\delta = 3, 5$ \cite{jia2021semantically} (Table \ref{tab:unmatched_results}). Although we outperform the baseline methods on the L2 distance metric (Dist in Table \ref{tab:unmatched_results}), we did not observe improvements in the pixel accuracy metrics. We do believe that our approach using VSA is capable of improving these metrics with a more suitable encoding method (as opposed to VGG) for images that have regions without contours or complex features (as is the case for some regions in Google Map images representing landscape or water). However, we leave study of encoding methods for VSA in computer vision applications to future research.

\subsection{Ablation Study}
\label{subsc:ablation}
\begin{figure*}[b!]
    \centering
    \includegraphics[width=\hsize]{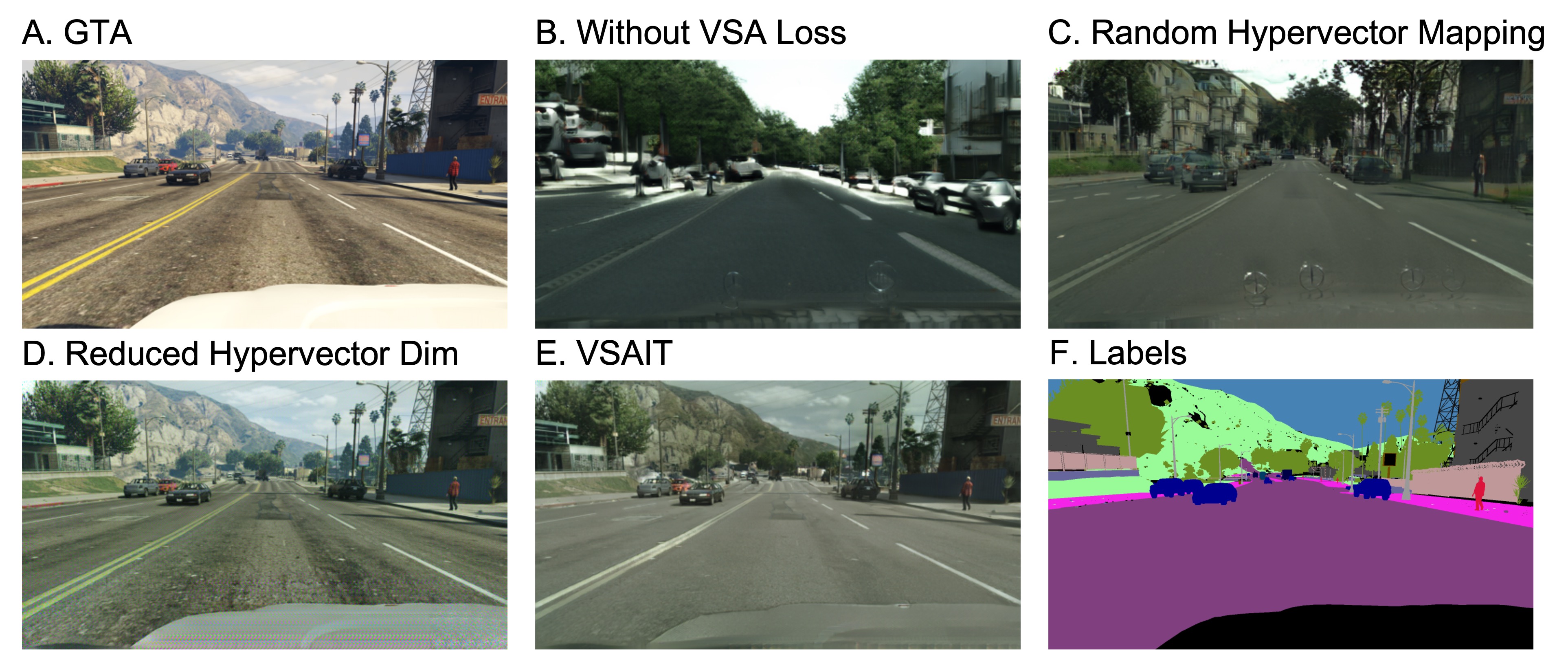}
    \caption{Ablation study for GTA to Cityscapes. (A) and (F) show the ground truth image and semantic labels for the GTA source image, respectively. (B) demonstrates the challenge of semantic flipping when using GAN-based methods without constraining the generator. (C) shows the importance of learning the hypervector mapping. (D) demonstrates the importance of high-dimensional space for this method. (E) shows the effect on image translation when incorporating all components of our approach.}
    \label{fig:ablation_study}
\end{figure*}

We demonstrate the effect of VSAIT by evaluating the contribution of the VSA-based cyclic loss (Eqn \ref{eqn:vsa_loss}), the learned hypervector mapping $u_{X \leftrightarrow Y}$, and the hypervector dimensionality. We use the GTA to Cityscapes task to evaluate these ablations on semantic segmentation performance metrics (Table \ref{tab:ablation_results}). We first demonstrate that by removing the VSA-based cyclic loss, the generator learns to translate images without preserving content, demonstrating the serious problem of semantic flipping (Figure \ref{fig:ablation_study}B). Next, we show the importance of the invertible mapping $u_{X \leftrightarrow Y}$ by generating a random hypervector mapping instead of learning the mapping adversarially (Eqn \ref{eqn:gan_loss}). When using a random hypervector instead of the learned mapping generated by $F$, the generator maintains global structure but since the invertible mapping does not recover the source hypervector (Eqn \ref{eqn:mapping_vector}) the local content is often changed (semantic flipping shown in Figure \ref{fig:ablation_study}C). Finally, we demonstrate the importance of using the VSA hypervector space by reducing the dimensionality of $\mathbb{V}$ from $4096$ to $128$. Reducing the dimensionality of the hyperspace from $4096$ to $128$ results in noisy image translations that seem to only reflect global changes in style. However, even using a low-dimensional hypervector VSAIT still outperforms most methods compared in Table \ref{tab:unmatched_results}.

\begin{table}[t!]
    \centering
    \caption{Evaluations on GTA to Cityscapes. Left: quantitative comparison for KID metric. *Mean values reported in~\cite{richter2021enhancing}. Right: ablation study task demonstrating the contributions for each component of VSAIT method.}
    \begin{adjustbox}{max width=\textwidth}
    \begin{tabular}{l|c}
        \hline
        Method & KID\\
        \hline
        CUT & 21.18* \\
        SRUNIT & 16.27 \\
        DRIT & 14.69 \\
        GcGAN & 12.32 \\
        EPE & 10.95* \\
        VSAIT (ours) & \textbf{8.74} \\
        \hline
    \end{tabular}
    \begin{tabular}{c|ccc}
        \hline
        Ablation & pxAcc & clsAcc & mIoU \\
        \hline
        w/o VSA Loss & 52.17 & 16.14 & 10.21 \\
        Random Hypervector Mapping & 65.75 & 29.32 & 17.53 \\
        Reduced Hypervector Dim & 66.42 & 40.34 & 25.04 \\
        \hline
        VSAIT (ours) & \textbf{76.48} & \textbf{45.33} & \textbf{30.89} \\
        \hline
    \end{tabular}
    
    \end{adjustbox}
    \label{tab:ablation_results}
\end{table}

\section{Conclusion}
In this paper, we address the semantic flipping problem in unpaired image translation using a novel approach based on VSA framework \cite{gayler2004vector,kanerva2009hyperdimensional}. We show important qualitative and quantitative improvements over previous methods that have attempted to address this problem, demonstrating that VSA can be used to invert image translations and ensure consistency with the source domain. Given its inherent versatility, we hope this work inspires future research using VSA in more computer vision applications.  \\

\noindent \textbf{Acknowledgments.}
We thank Mihir Jain, Shingo Takagi, Patrick Rodriguez, Sarah Watson, Zijian He, Peizhao Zhang, and Tao Xu for their helpful feedback.

\printbibliography

\clearpage
\appendix

\begin{refsection}
\section{Additional Implementation Details}

In order to encode image patches into the VSA hyperspace, we extract features using VGG19 \cite{simonyan2014very} pre-trained on ImageNet \cite{deng2009imagenet}. We select multiple layers from VGG19 and concatenate features that share the same receptive field. For all experiments, we extract features from multiple convolutional layers in VGG19. For GTA to Cityscapes, we concatenate features for patch sizes $16 \times 16$, $8 \times 8$, $4 \times 4$, $2 \times 2$, and, $1 \times 1$. This means that extracted features within each patch are concatenated into a single feature vector, which is reduced in dimensionality using locality sensitive hashing as described in Section 3.3 of the main paper. For all experiments, we used a hypervector dimensionality of $4096$, which is the dimensionality of the input to the discriminator and mapping networks. We then used 1024 channels in all layers except the last (1 channel) for the discriminator, and 4096 channels for both layers in the mapping network. 

For Google Map to Aerial Photo, we use the same patch sizes as for GTA to Cityscapes. However, for Aerial Photo to Google Map, we empirically found that using ``dilated'' patches gave better performance. For this experiment, we use features extracted from four convolutional layers with patch sizes $32 \times 32$, $16 \times 16$, $8 \times 8$, and $4 \times 4$, where each patch has spaces between locations in the extracted feature maps (i.e., equivalent to a convolutional dilation factor of $3$).

For all experiments, we trained using $256 \times 256$ images (as was done in the baseline methods). In the case of GTA to Cityscapes, we first resize the images to a height and width of $256 \times 512$ followed by random cropping to $256 \times 256$. For the Google Map and Aerial Photo datasets, images are resized from $600 \times 600$ to $256 \times 256$. Note that we do not report results for EPE in Table 1 of the main paper as we are not able to replicate their method using $256 \times 256$ image resolution for GTA to Cityscapes and it is not feasible to use their method for datasets without G-buffers. As was done in \cite{jia2021semantically}, we train for 20 epochs for GTA to Cityscapes and 400 epochs for Google Map to Aerial Photo and Aerial Photo to Google Map experiments. We multiply the learning rates (defined in Section 4.1 of the main paper) by a factor of $0.5$ every $100$K iterations.

\section{Difference between VSA-Based Cyclic Loss and Perceptual Loss}
Many image-to-image translation methods have included a so-called ``perceptual loss'' to minimize the distance between translated and source images in VGG feature space. Although our VSA-based cyclic loss (Eqn 6) may seem similar to the perceptual loss, our method is distinct: we minimize the distance between source hypervectors and translated hypervectors that are mapped back from the target to source domain, whereas methods using a perceptual loss directly minimize the distance between source and translated features. Furthermore, as noted previously, encoding these source and translated features into the VSA hyperspace gives greater assurance that different semantic content will be almost orthogonal to each other. 

\section{Additional Ablation Studies}
As indicated above, we changed the patch sizes used during feature extraction across different experiments (i.e., ``dilated'' patches in Aerial Photo to Google Map). We therefore, also looked at the effect of patch size within a specific experiment. In addition to the $16 \times 16$ patch size used in the GTA to Cityscapes experiment, we also tested $32 \times 32$ and $8 \times 8$ patch sizes. Compared to the results in Table 1 (mIoU 30.89), we observed mIoU of 32.29 and 31.47 for 32 and 8 patch sizes, respectively.

Furthermore, we conducted an ablation of the weight ($\lambda$) for our VSA-based cyclic loss, showing that $\lambda < 5$ results in semantic flipping, which is reflected in lower mIoU for the GTA to Cityscapes experiment. Specifically, we observed lower mIoU for smaller (21.55 and 26.49 for $\lambda = 1, 2$) compared to larger $\lambda$ (29.13, 30.89, 29.78 for $\lambda = 5, 10, 20$).

Finally, we also tested the effect of the hypervector dimensionality. As demonstrated in~\cite{neubert2019introduction}, the hypervector dimensionality should be at least greater than 700 (the probability of two randomly sampled vectors being orthogonal approaches 1). We tested 512 and 2048 dimensional hypervectors, demonstrating worse performance for 512 compared to 2048 (29.35 vs. 30.87 mIoU for GTA to Cityscapes, respectively).

\section{Hypervector Adversarial Loss}
Since our hypervector adversarial loss (Eqn 5) has two negative targets and one positive target, it's possible that this could lead to an imbalance in training. For the current experiments, we have tested using a balanced weighting of the loss terms and did not empirically observe differences in stability or performance. However, this may be an important consideration for future research with other datasets.

\section{Quantitative Comparison to Additional Methods for GTA to Cityscapes}
In order to compare our method against other relevant methods that have also been applied to GTA to Cityscapes, we trained semantic segmentation networks using our translated GTA images as done in other methods. In these comparisons we used ResNet-50 as the feature extractor rather than VGG19. Specifically, we compared our method against CyCADA~\cite{hoffman2018cycada} and Fourier Domain Adaptation (FDA)~\cite{yang2020fda}. CyCADA~\cite{hoffman2018cycada} is a method that uses the cycle-consistency approach from CycleGAN~\cite{zhu2017unpaired} in order to ensure semantic consistency between segmentation predictions before and and after image translation. Unlike GAN-based methods, FDA~\cite{yang2020fda} computes the Fourier Transform of source and target images, and replaces the amplitudes of low-frequency features in the source images with those from the target images. We observed comparable performance to these approaches (45.7 mIoU for VSAIT vs. 42.7 in CyCADA~\cite{hoffman2018cycada} and 44.6 in FDA~\cite{yang2020fda}).

We further compared our results to those from~\cite{wang2022learning}, which is a recent video-to-video translation technique that aims to prevent semantic flipping via pseudo-supervision with optical flow from sequential video frames. By using a checkpoint from the authors to evaluate semantic segmentation performance (GTA to Cityscapes, Table 1), we observed lower performance compared to our method (pixel accuracy: 66.04, class accuracy: 38.42, and mIoU: 24.53).

\section{VSA Hypervector Encoding}
As noted in the main paper, VGG19 works well for most natural image domains; however, it may not be optimal for encoding semantic segmentation masks or similar types of images (e.g., Google Map images). For example, using VSAIT to translate semantic masks to Cityscapes images using the same sub-sampling method used in \cite{jia2021semantically}, we observed worse performance compared to that reported in \cite{jia2021semantically} (pixel accuracy: 69.28, class accuracy: 31.37, mIoU: 21.59). Therefore, an important future direction for this research is to understand and improve encoding methods when using VSA for computer vision applications.

\section{Additional Visual Examples}
\subsection{GTA $\to$ Cityscapes}

\begin{figure*}[h!]
    \centering
    \includegraphics[width=\hsize]{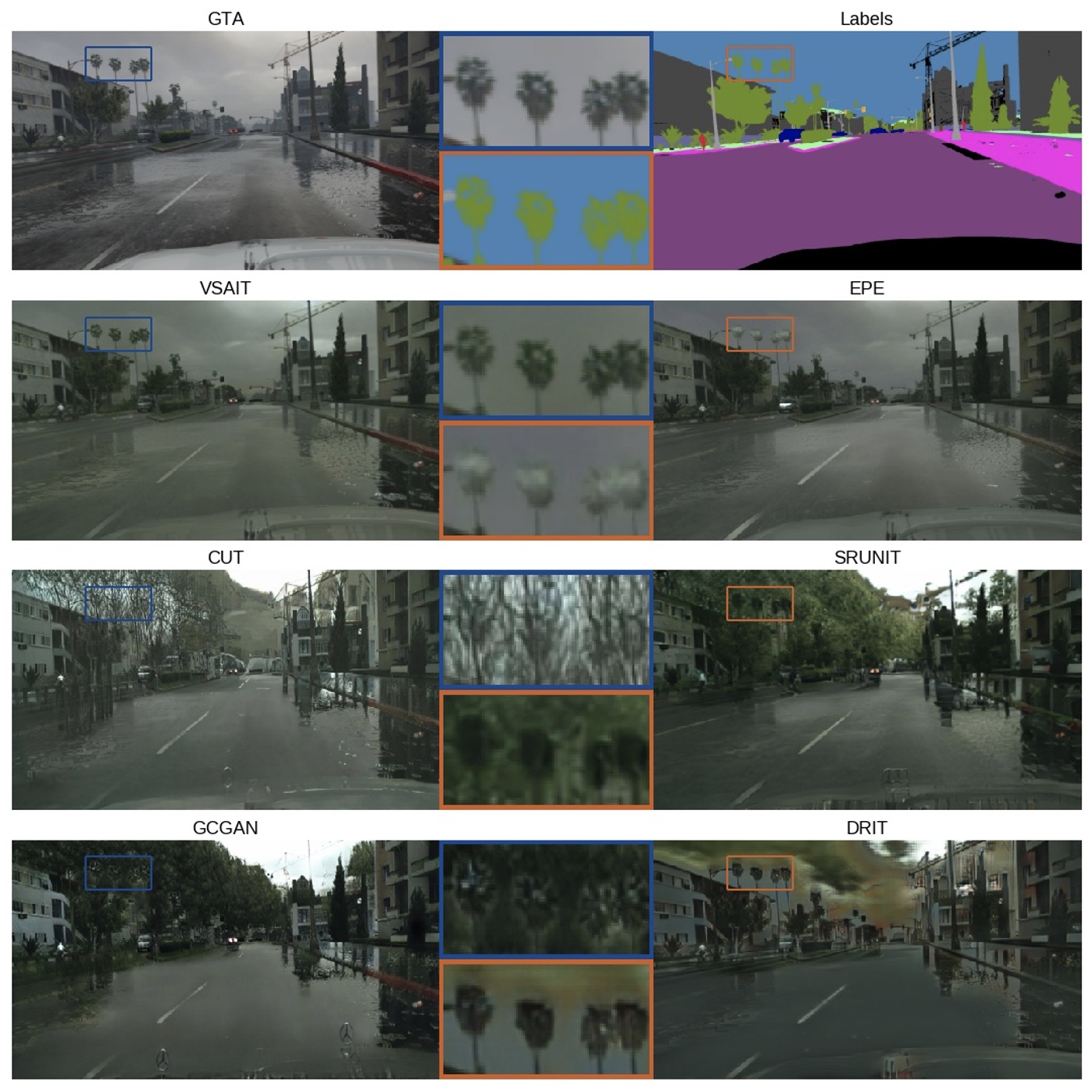}
    \caption{Examples of semantic flipping for baseline methods in GTA to Cityscapes experiment.}
    \label{fig:gta2cityscapes_1}
\end{figure*}

\begin{figure*}[h!]
    \centering
    \includegraphics[width=\hsize]{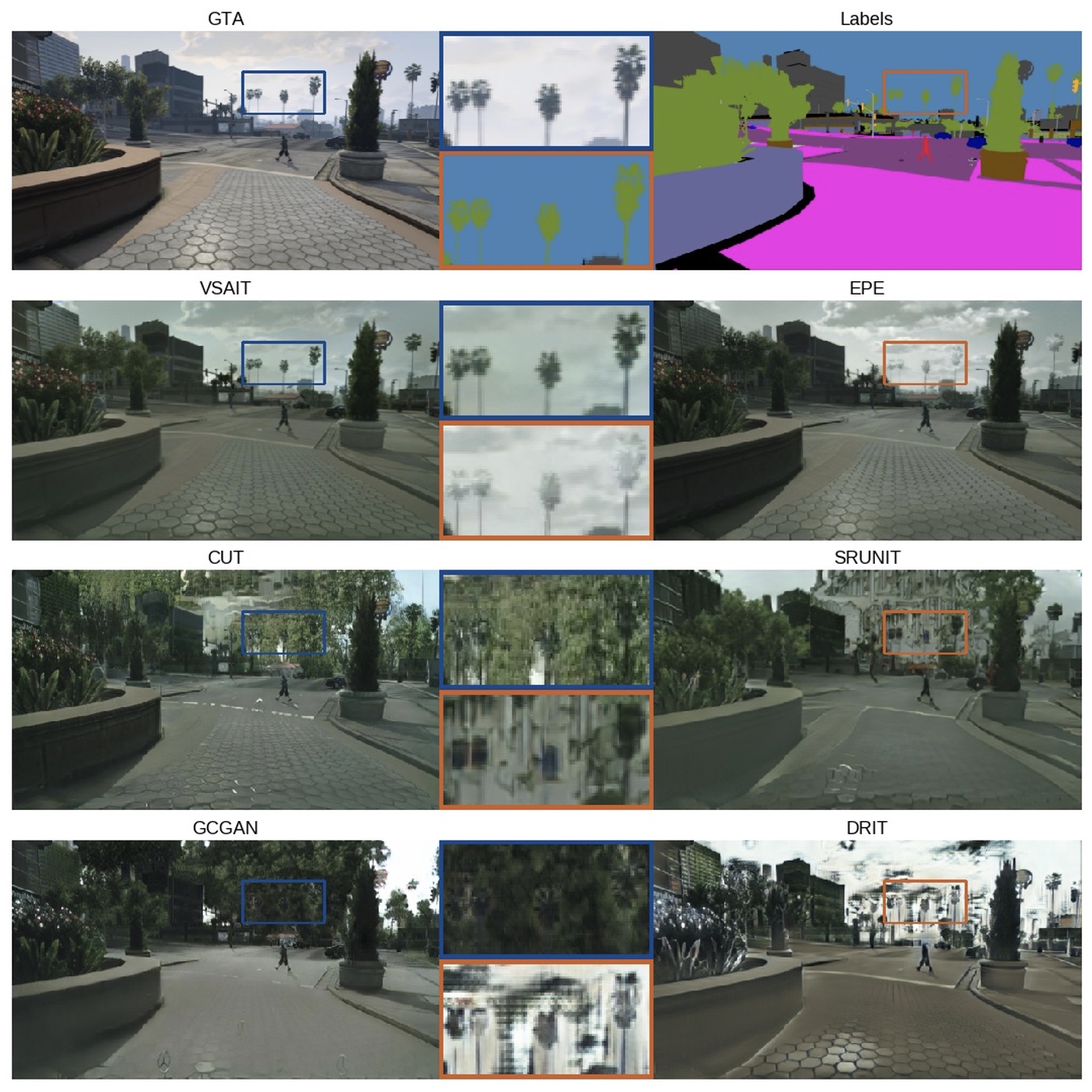}
    \caption{Examples of semantic flipping for baseline methods in GTA to Cityscapes experiment.}
    \label{fig:gta2cityscapes_3}
\end{figure*}

In addition to the comparisons shown in the main paper, we provide more qualitative examples showing our method relative to the baseline methods for GTA to Cityscapes. As described in the main paper, GTA and Cityscapes have semantics that are naturally different. Most notably, Cityscapes has more vegetation in the upper half of the image whereas GTA has more sky (see also Figure 1 from the main paper). Therefore, many GAN-based image translation methods hallucinate trees in the sky as shown in Figures \ref{fig:gta2cityscapes_1} and \ref{fig:gta2cityscapes_3}. More interestingly, EPE does not hallucinate trees but instead removes palm trees from images. This is likely due to the fact that palm trees do not appear in Cityscapes. On the other hand, our method correctly retains the palm trees without hallucinating more trees in the sky.

\begin{figure*}[h!]
    \centering
    \includegraphics[width=\hsize]{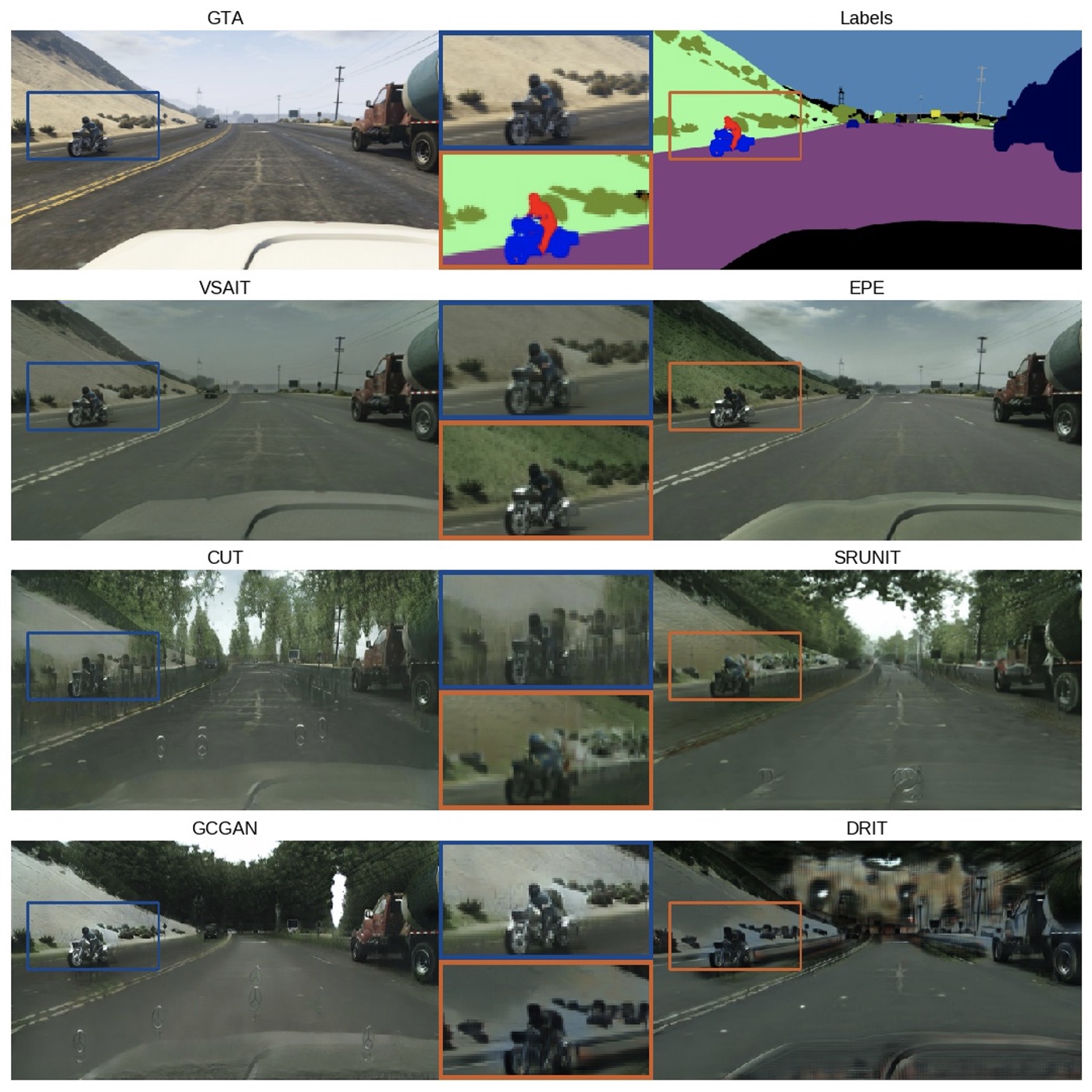}
    \caption{Examples of semantic flipping for baseline methods in GTA to Cityscapes experiment.}
    \label{fig:gta2cityscapes_4}
\end{figure*}

Figure \ref{fig:gta2cityscapes_4} provides a similar example of the typical hallucinations that occur when translating GTA images with broad sky regions. However, notice the sandy hill on the left side of the image, which we highlight in the figure. As shown in the top-right panel, the semantic segmentation label for this region is the same for both the grassy and sandy portions of the hill. Since EPE uses these labels as part of its task-level consistency loss, it generates the same feature (i.e., grass) across the region, irrespective of the underlying specific content differences.

\clearpage

\clearpage

\subsection{Google Map $\to$ Aerial Photo}

\begin{figure*}[h!]
    \centering
    \includegraphics[width=\hsize]{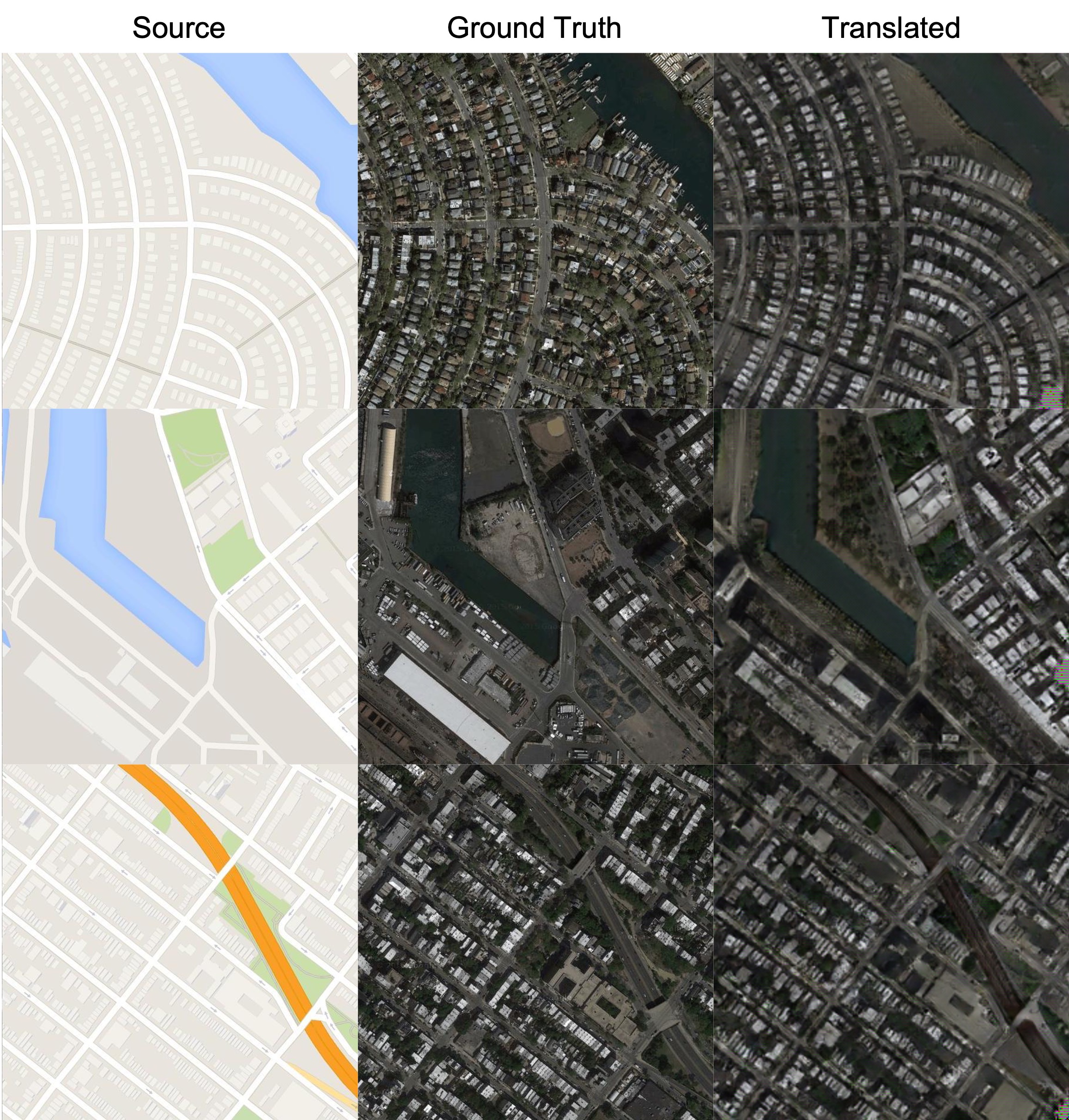}
    \caption{Example image translations using VSAIT for the Google Map to Aerial Photo experiment.}
    \label{fig:map2photo_1}
\end{figure*}

\begin{figure*}[h!]
    \centering
    \includegraphics[width=\hsize]{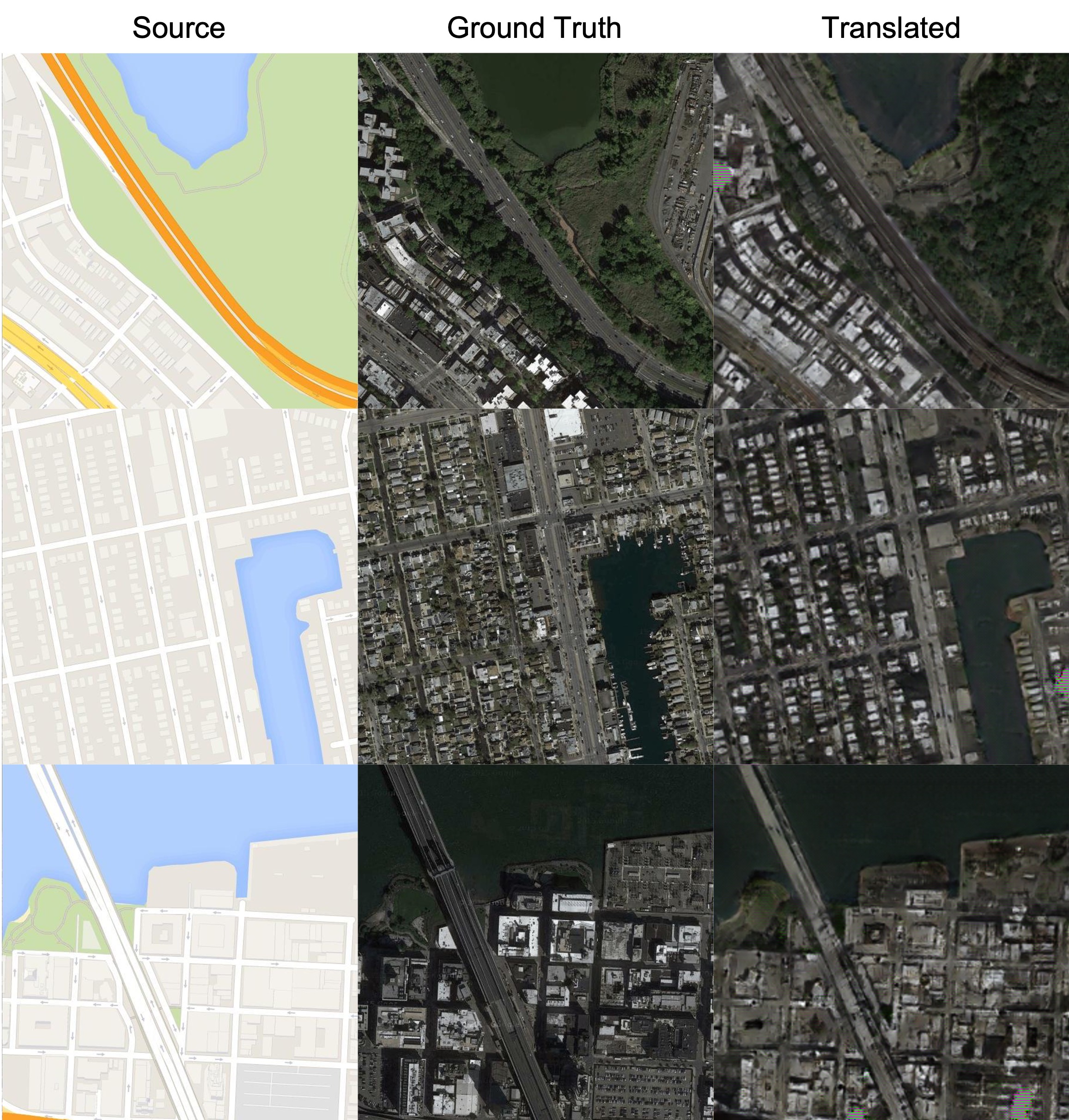}
    \caption{Example image translations using VSAIT for the Google Map to Aerial Photo experiment.}
    \label{fig:map2photo_2}
\end{figure*}

\begin{figure*}[h!]
    \centering
    \includegraphics[width=\hsize]{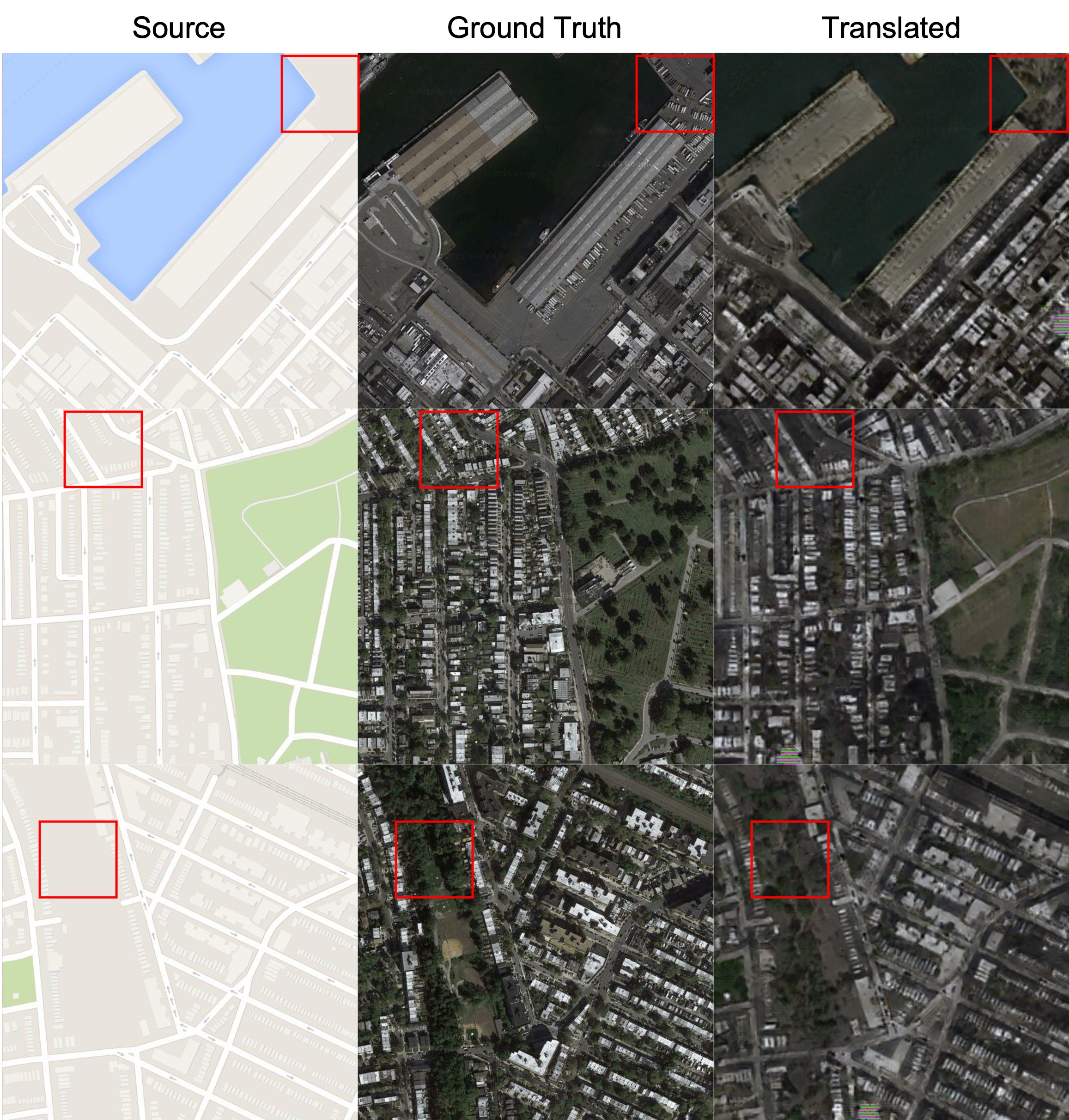}
    \caption{Example image translations using VSAIT for the Google Map to Aerial Photo experiment.}
    \label{fig:map2photo_3}
\end{figure*}

As described in Section 4.5 of the main paper, we sub-sample the Google Map and Aerial Photo datasets (following \cite{jia2021semantically}) to obtain unpaired source and target datasets that differ in their semantic statistics. As a result, the sub-sampled datasets differ in the proportion of image area containing land versus water. Therefore, similar to the GTA to Cityscapes experiment, we would expect hallucinations of houses or trees in the water. We demonstrate across multiple examples in Figures \ref{fig:map2photo_1} and \ref{fig:map2photo_2} that our method does not exhibit semantic flipping. It's possible that some methods may rely on a one-to-one mapping using the RGB values in the map images (e.g., blue for water) that could reduce the diversity of translations overall. However, as shown in Figure \ref{fig:map2photo_3}, we observe diverse translations for the same RGB values across different map images. Specifically, notice the translations associated with gray pixels in each source image as highlighted in the figure.

\clearpage

\subsection{Aerial Photo $\to$ Google Map}

\begin{figure*}[b!]
    \centering
    \includegraphics[width=\hsize]{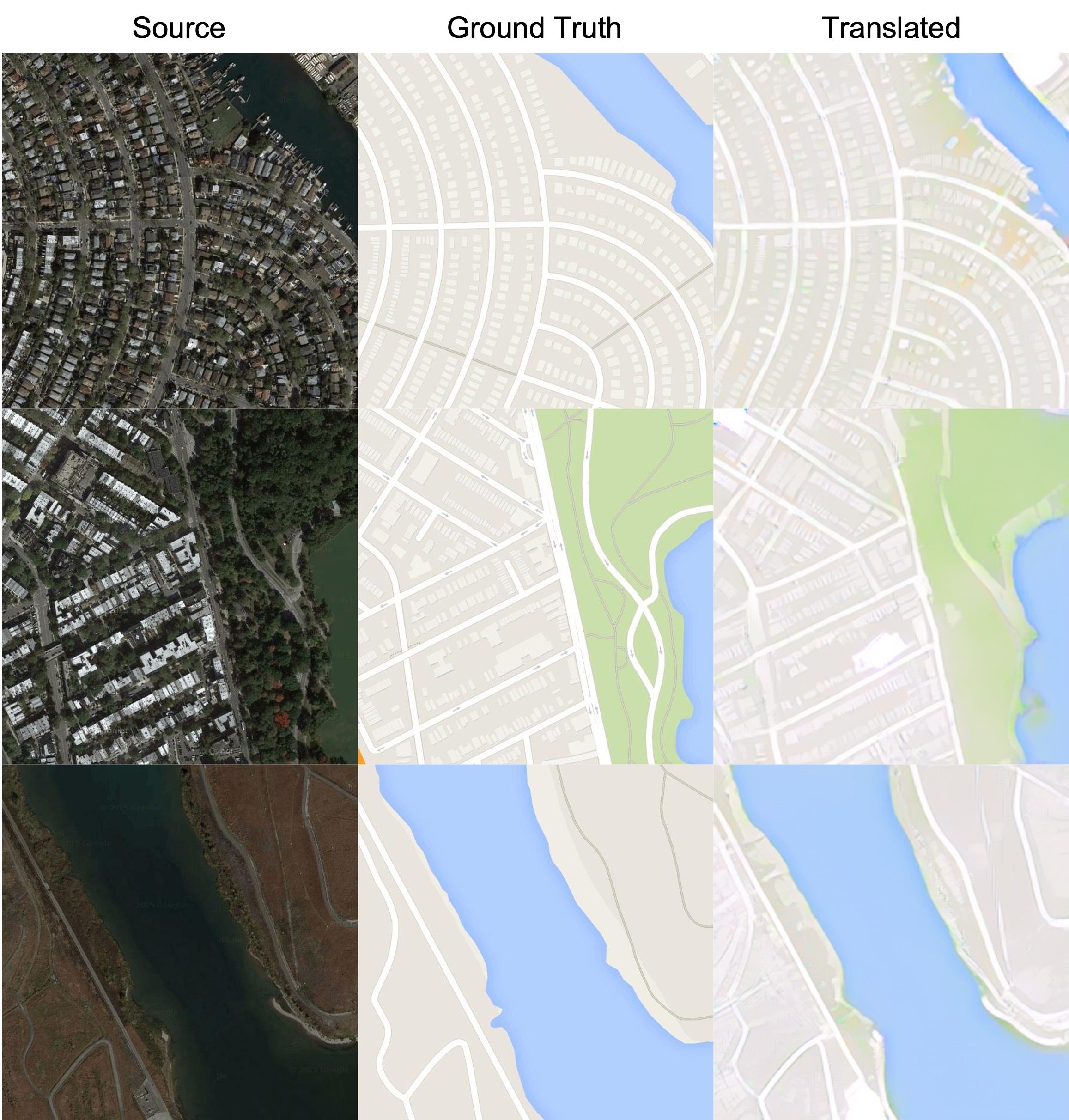}
    \caption{Example image translations using VSAIT for the  Aerial Photo to Google Map experiment.}
    \label{fig:photo2map_1}
\end{figure*}

\begin{figure*}[b!]
    \centering
    \includegraphics[width=\hsize]{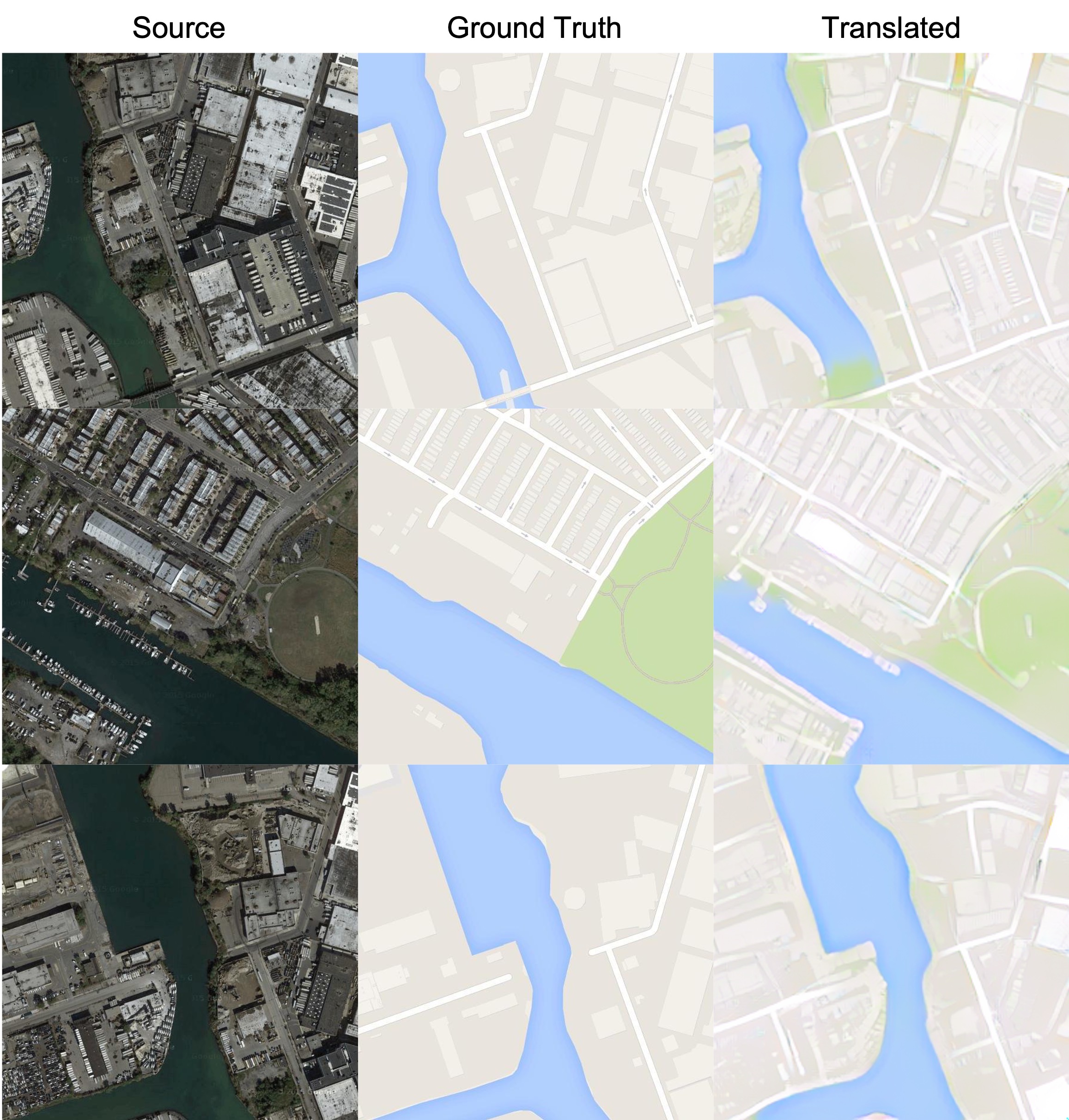}
    \caption{Example image translations using VSAIT for the  Aerial Photo to Google Map experiment.}
    \label{fig:photo2map_2}
\end{figure*}

\begin{figure*}[b!]
    \centering
    \includegraphics[width=\hsize]{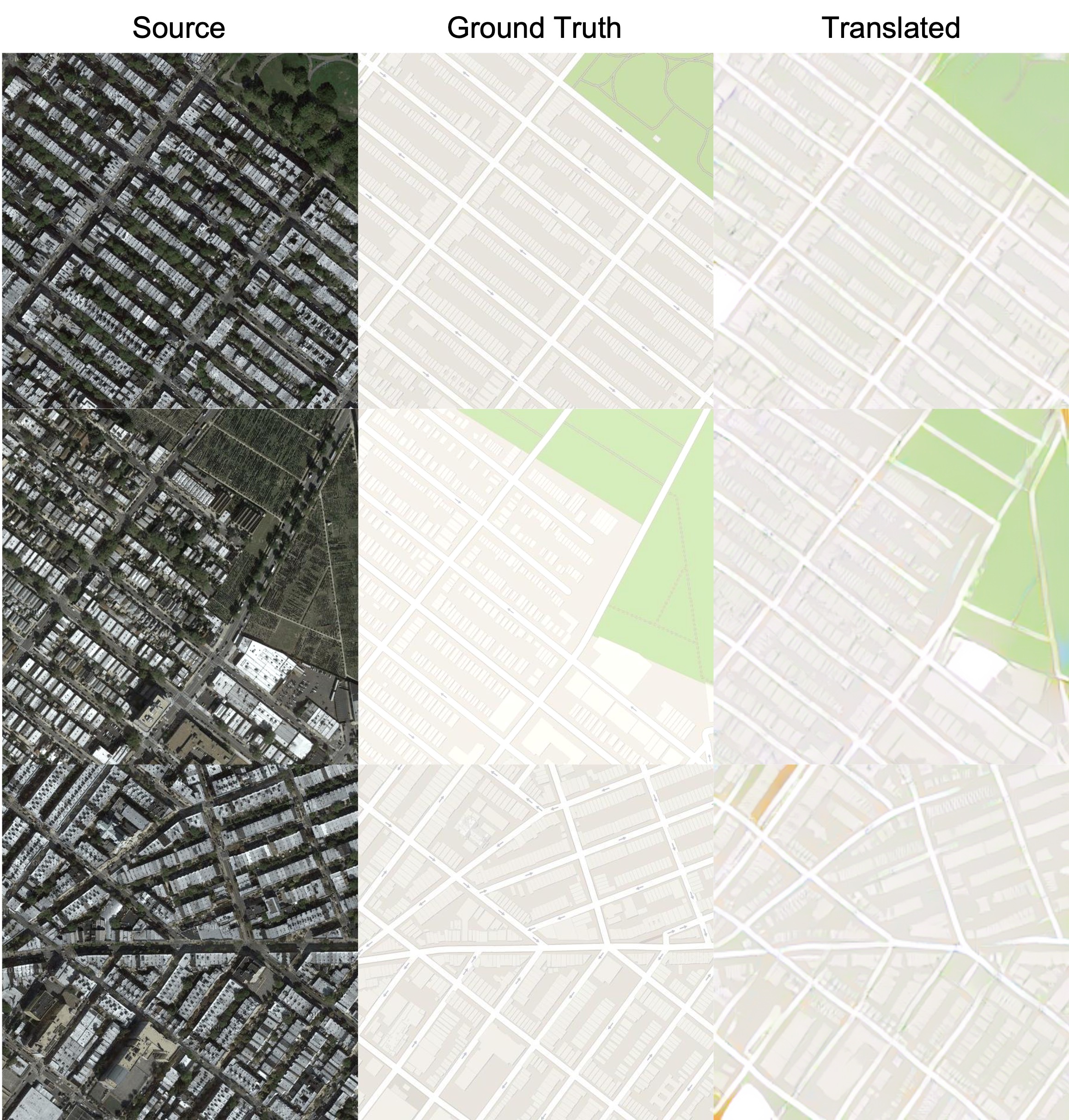}
    \caption{Example image translations using VSAIT for the  Aerial Photo to Google Map experiment.}
    \label{fig:photo2map_4}
\end{figure*}

Finally, we demonstrate examples of our image translations from the Aerial Photo to Google Map experiment. As mentioned previously, we use the same sub-sampling procedure to obtain datasets with unmatched semantic statistics. Since the sub-sampled Google Map dataset has more regions with water relative to the sub-sampled Aerial Photo dataset, we would expect that GAN-based methods might suffer from semantic flipping by hallucinating water. However, as shown in Figures \ref{fig:photo2map_1} and \ref{fig:photo2map_2}, our method learns the correct mapping between water and does not exhibit semantic flipping in scenes without water (Figure \ref{fig:photo2map_4}).

\clearpage

\subsection{Visual Comparison to Baseline Methods}

\begin{figure*}[h!]
    \centering
    \includegraphics[width=\hsize]{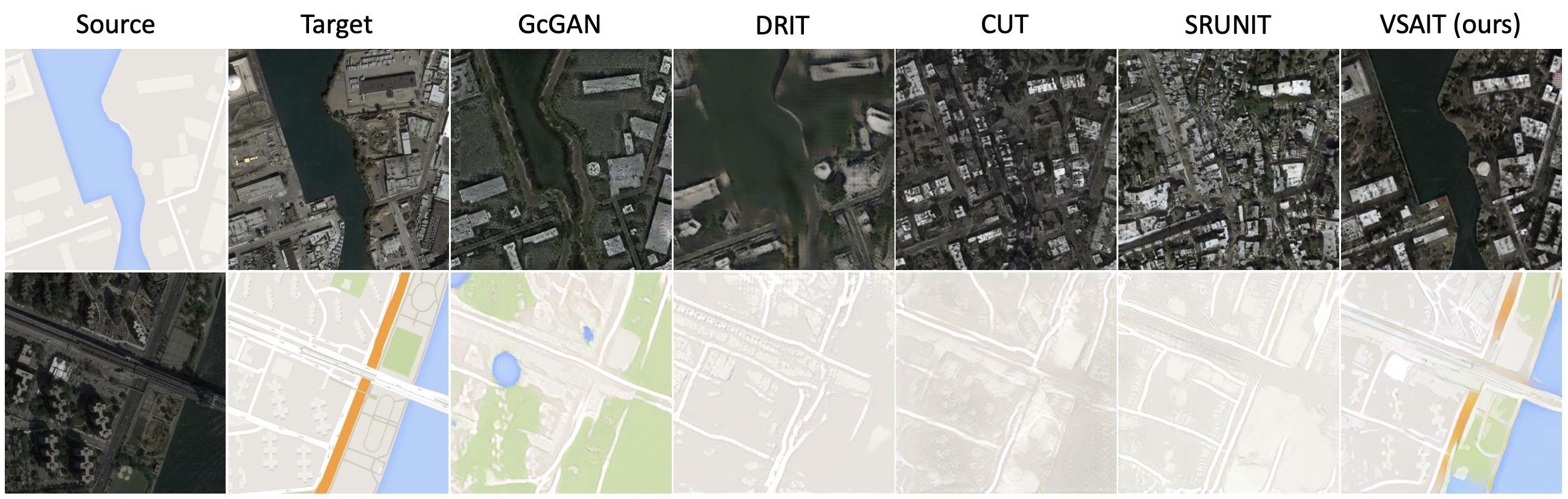}
    \caption{Example image translations for VSAIT and baseline methods for Aerial Photo to Google Map and Google Map to Aerial Photo experiments. Note that EPE cannot be evaluated for these experiments.}
    \label{fig:baseline_comp}
\end{figure*}

We additionally demonstrate the comparison between VSAIT and baseline methods for the Aerial Photo to Google Map and Google Map to Aerial Photo experiments in Figure \ref{fig:baseline_comp}. As shown in the figure, for these more difficult experiments where the datasets are specifically sub-sampled to reduce content overlap between domains, most of the baseline methods exhibit considerable semantic flipping (e.g., water to land).

\printbibliography[heading=subbibliography]
\end{refsection}

\end{document}